\documentclass{article}
\usepackage{microtype}
\usepackage{graphicx}
\usepackage{subcaption}
\usepackage{booktabs} 
\usepackage{hyperref}

\usepackage[accepted]{icml2026}

\usepackage{amsmath}
\usepackage{amssymb}
\usepackage{mathtools}
\usepackage{amsthm}

\usepackage[capitalize,noabbrev]{cleveref}

\theoremstyle{plain}

\theoremstyle{definition}

\theoremstyle{remark}

\usepackage[textsize=tiny]{todonotes}

\icmltitlerunning{Submission and Formatting Instructions for ICML 2026}

\begin{document}

\twocolumn[
  \icmltitle{Frequency-Decomposed INR \\for NIR-Assisted Low-Light RGB Image Denoising}



  \icmlsetsymbol{equal}{*}

  \begin{icmlauthorlist}
    \icmlauthor{Ligen Shi}{slg}
    \icmlauthor{Zengyu Pang}{pzy}
    \icmlauthor{Chang Liu}{jss}
    \icmlauthor{Shuchen Sun}{jss}
    \icmlauthor{Jun Qiu}{jss}
  \end{icmlauthorlist}
  
  \icmlaffiliation{slg}{College of Computer Science (College of Software), Inner Mongolia University, Hohhot, 010021, China}
   
  \icmlaffiliation{pzy}{School of Mathematics and Statistics, Wuhan University, Wuhan, 430072, China}
  
  \icmlaffiliation{jss}{Institute of Computational Imaging, Beijing Information Science and Technology University, Beijing
  	102206, China}
  	
  \vskip 0.3in
]



\printAffiliationsAndNotice{}  

\begin{abstract}
Addressing the issues of severe noise and high-frequency structural degradation in visible images under low-light conditions, this paper proposes a Near-Infrared (NIR) aided low-light image restoration method based on Frequency-Decoupled Implicit Neural Representation (FD-INR). Based on the statistical prior of RGB-NIR cross-modal frequency correlations — specifically that low-frequency RGB signals are more reliable, whereas high-frequency NIR signals exhibit higher correlation — we explicitly decompose images into distinct frequency components via multi-scale wavelet transforms and construct a dual-branch implicit neural representation framework. Within this framework, we design a cross-modal differentiated frequency supervision mechanism, leveraging low-light RGB to guide the reconstruction of low-frequency luminance and color, and utilizing high-SNR NIR signals to constrain the generation of high-frequency texture details, thereby achieving complementary advantages in the frequency domain. Furthermore, an uncertainty-based adaptive weighting loss function is introduced to automatically balance the contributions of different frequency tasks, solving the problems of color distortion and artifacts caused by rigid fusion in the spatial domain common in traditional methods. Experimental results demonstrate that FD-INR not only effectively restores image luminance consistency and structural details but also, benefitting from its implicit continuous representation, outperforms existing methods in arbitrary-resolution reconstruction tasks, significantly enhancing the reliability of low-light perception.
\end{abstract}

\section{Introduction} \label{sec:introduction}

With the rapid advancement of Unmanned Aerial Vehicle (UAV) technology, autonomous perception in all-weather environments has become a focal point in both academia and industry. However, UAVs often encounter severe lighting challenges when operating at dawn, dusk, in forest shadows, or at night. Due to strict constraints on payload weight and power consumption, these platforms typically use small-scale imaging sensors, resulting in insufficient photon counts under low-light conditions \cite{wei2020physics, li2021low, chen2018learning}. Consequently, captured RGB images are heavily degraded by quantum and electronic noise, which masks high-frequency textures and causes color bias and contrast degradation. This decline in low-level visual quality weakens the robustness of UAVs in downstream tasks such as scene understanding, object detection, and 3D reconstruction \cite{ye2021darklighter, li2022all}.

To overcome the physical limits of single-modal visible light sensors, Near-Infrared (NIR) assisted imaging has emerged as a promising solution \cite{yan2013cross, wan2022purifying, wang2025complementary}. Due to its longer wavelength, NIR signals exhibit higher penetration and quantum efficiency in low light, providing structural representations with superior signal-to-noise ratios (SNR). However, cross-modal fusion between RGB and NIR is challenging due to physical inconsistencies in their imaging mechanisms. First, varying material reflectances across spectral bands lead to distinct spatial luminance distributions (e.g., vegetation vs. fabric). Second, RGB images exhibit frequency-selective degradation in low light: while low-frequency components (global contours) remain relatively reliable, high-frequency components (local textures) are often dominated by noise.

Existing research primarily involves single-modal enhancement (e.g., Retinex-based methods or deep CNNs), which often faces a trade-off between noise suppression and detail preservation, frequently resulting in over-smoothing or artifacts \cite{wei2018deep, zhang2017beyond}. Current multi-modal fusion strategies generally follow two directions. The first category performs direct fusion in the spatial domain \cite{li2018densefuse, xu2022model, sheng2022frequency}. These methods rely on neural network fitting but often neglect inconsistent spatial dependencies and non-linear luminance differences between modalities \cite{ma2019fusiongan}, leading to color shifts and artifacts. The second category employs decomposition or attention mechanisms to enhance shared features while suppressing inconsistent ones \cite{yan2013cross, kim2021deformable, denger2020deep}. While this reduces interference, it often discards modality-specific information—such as high-frequency NIR textures or unique RGB color data—yielding suboptimal results. Furthermore, most existing spatial or decomposition methods rely on discrete, grid-based representations and lack quantitative modeling of frequency reliability. This discrete sampling limits the model's ability to adaptively handle varying noise levels and constrains fine-grained representation at continuous resolutions.

To address these limitations, we propose FD-INR, an Implicit Neural Representation framework based on explicit frequency decomposition for RGB-NIR collaborative reconstruction. Unlike previous discrete modeling approaches, we represent the image field as an implicit continuous function, providing a coordinate-aware unified feature space. Our approach is motivated by the observation of cross-field frequency correlations: RGB low-frequencies correlate strongly with the ground truth, while NIR high-frequencies characterize scene structures more accurately than noise-polluted RGB. Based on this prior, we utilize Discrete Wavelet Transform (DWT) to construct a dual-branch implicit learning mechanism: 
\begin{enumerate}
	\item \textbf{Low-frequency Consistency Branch}: Uses the low-light RGB image as the primary supervision to ensure the reconstructed image maintains natural color and global luminance. \item \textbf{High-frequency Structure Enhancement Branch}: Extracts reliable gradient information from the NIR image to compensate for and reconstruct the RGB high-frequency details masked by noise. 
\end{enumerate} 

Additionally, to adaptively balance the contributions of different frequency components under varying noise levels, we introduce a weighted loss function based on uncertainty learning. By modeling homoscedastic uncertainty as learnable parameters, FD-INR dynamically adjusts the weights of each frequency branch during training. We further employ the Muon optimizer to improve the convergence efficiency of the INR in high-dimensional coordinate mapping, maximizing detail recovery while maintaining color fidelity.

The primary contributions of this work are as follows: \begin{itemize} \item We propose FD-INR, an implicit neural representation method based on wavelet decoupling. This method integrates the physical prior of "RGB-constrained low frequencies and NIR-enhanced high frequencies" into a continuous function space. Through explicit frequency partitioning, it addresses modal conflict and feature aliasing inherent in traditional spatial fusion, enabling high-quality RGB-NIR reconstruction at arbitrary resolutions. \item We design an adaptive weight allocation mechanism based on homoscedastic uncertainty and incorporate an orthogonal optimizer (Muon) to enhance the learning efficiency of the continuous representation, achieving a balance between color preservation and detail restoration. \item Extensive experiments demonstrate that FD-INR achieves state-of-the-art (SOTA) performance on multiple low-light benchmarks and exhibits strong generalization in real-world extreme low-light scenarios, significantly improving the robustness of downstream tasks such as object detection. \end{itemize}

\section{Related Work} \label{sec:related_work}

This section reviews related research from three perspectives. First, we discuss single-modal low-light enhancement and its limitations in high-frequency restoration. Second, we analyze NIR-assisted enhancement methods and the challenges posed by modal inconsistency. Finally, we summarize progress in Implicit Neural Representations (INR) and articulate how our proposed frequency decoupling and continuous representation address the bottlenecks of existing fusion methods.

\subsection{Low-Light Image Restoration} Early low-light enhancement research primarily followed the physics-driven Retinex Theory \cite{land1977retinex, Land71}, which achieves brightness enhancement by modeling illumination and reflectance. Representative works include multi-scale Retinex (MSR) \cite{jobson1997multiscale} and the optimization-constrained LIME \cite{guo2016lime}. However, these methods often struggle to distinguish noise amplified by high gains from genuine reflective details when processing images with extremely low signal-to-noise ratios (SNR).

With the rise of deep learning, data-driven methods have become dominant. Early works such as LLNet \cite{lore2017llnet} demonstrated the potential of autoencoders in denoising, while See-in-the-Dark (SID) \cite{chen2018learning} bypassed the limitations of traditional ISPs by directly processing RAW data. To reduce reliance on paired data, Zero-DCE \cite{guo2020zero} and EnlightenGAN \cite{jiang2021enlightengan} introduced unsupervised curve mapping and adversarial generative networks, respectively. Subsequently, researchers explored complex architectures, such as Transformer-based Restormer \cite{zamir2022restormer} and multi-scale diffusion models \cite{jiang2024lightendiffusion}, to improve restoration quality in complex degradation scenarios. However, these methods face a fundamental challenge: the irreversibility of information loss. Under extremely weak illumination, high-frequency signals captured by RGB sensors are almost entirely masked by thermal and shot noise, causing models to produce "hallucinated artifacts" or unnatural smoothing when attempting to restore edges. This inherent limitation of single-modal restoration has motivated researchers to explore multi-modal information complementarity.

\subsection{NIR-Assisted Image Enhancement} The near-infrared (NIR) band (700nm-1100nm) can capture high-SNR texture details without relying on strong visible light sources due to its physical properties. The use of NIR to assist visible light imaging dates back to traditional computational photography. Petschnigg et al. \cite{petschnigg2004digital} introduced flash/no-flash image pair processing, laying the foundation for cross-modal enhancement. Subsequently, the Guided Filter proposed by He et al. \cite{he2012guided} and the Scale Map mechanism by Yan et al. \cite{yan2013cross} utilized gradient transfer to guide RGB structural restoration via NIR high-frequency edges, becoming classic paradigms in the field.

With the prevalence of deep learning, researchers have utilized convolutional neural networks (CNNs) to mine deep correlations in cross-modal data. Early explorations like CUNet \cite{deng2020deep} used convolutional sparse coding to extract common features. To address non-linear differences between modalities, MNNet \cite{xu2022model} proposed an observation model explicitly considering the differences between target and guide images, while DVN \cite{jin2022darkvisionnet} injected structural inconsistency priors into deep networks. Furthermore, for flexible feature fusion, NAID \cite{xu2024nir} designed a Selective Fusion Module (SFM) that can be integrated into various denoising networks. Beyond spatial domain methods, researchers have also focused on cross-modal characteristics in the frequency domain. FGDNet \cite{sheng2022frequency} and SANet \cite{sheng2023structure} attempted to fuse features in the frequency domain to avoid spatial alignment difficulties. Recently, FCENet \cite{wang2025complementary} revealed cross-field frequency correlation priors from a statistical perspective, proving that cross-modal consistency is significantly higher in high-frequency bands than in low-frequency bands, providing theoretical support for resolving representation conflicts. Cross-modal alignment and deployment flexibility have also gained prominence. Kim et al. \cite{kim2025pixel} recently provided a sub-pixel aligned RGB-NIR dataset and demonstrated the potential for training-free inference. This move away from large-scale pre-training dependencies toward universal, instance-specific enhancement motivates our research direction.

Existing NIR-assisted methods still struggle to simultaneously maintain color accuracy, detail preservation, and artifact suppression under extreme low light. Traditional methods are prone to gradient artifacts, while deep learning methods often ignore the physical differences between frequency components, leading to color distortion. Inspired by FCENet \cite{wang2025complementary}, our proposed FD-INR extends frequency partitioning to the continuous implicit representation space. Through explicit high- and low-frequency decoupling branches, it preserves NIR high-frequency textures while avoiding color contamination caused by physical inconsistency.

\subsection{Implicit Neural Representation} 
Implicit Neural Representation (INR) \cite{xie2022neural, essakine2024we} revolutionizes the discrete grid-based paradigm by modeling signals as continuous mapping functions from coordinates to pixel attributes. Unlike discrete representations, INR can theoretically capture continuous details at arbitrary resolutions. The core lies in using Multi-Layer Perceptrons (MLPs) with activation functions to fit the non-linear mapping from spatial coordinates to signal values. DeepSDF \cite{park2019deepsdf} and NeRF \cite{mildenhall2021nerf} demonstrated this paradigm's capability in 3D geometry and radiance fields. However, early fully connected networks faced bottlenecks in computational efficiency and convergence. To address frequency bias and efficiency, SIREN \cite{sitzmann2020implicit} introduced periodic activation functions, while recent studies \cite{muller2022instant, chen2022tensorf, xie2023diner} shifted toward structured feature grids, combining feature lookups with lightweight MLPs. To overcome slow training, recent orthogonal gradient-based optimizers, such as Muon \cite{mcginnis2025optimizing}, have also shown potential.

For 2D image tasks, LIIF \cite{chen2021learning} proposed local implicit functions to achieve resolution-independent representation based on local latent codes. Subsequent works introduced multi-scale mechanisms; for instance, IPE \cite{barron2021mip} addressed anti-aliasing, and LTE \cite{lee2022local} enhanced spectral fidelity by estimating local textures in the Fourier domain.

\begin{figure*}[t] 
	\centering
	\includegraphics[width=\textwidth]{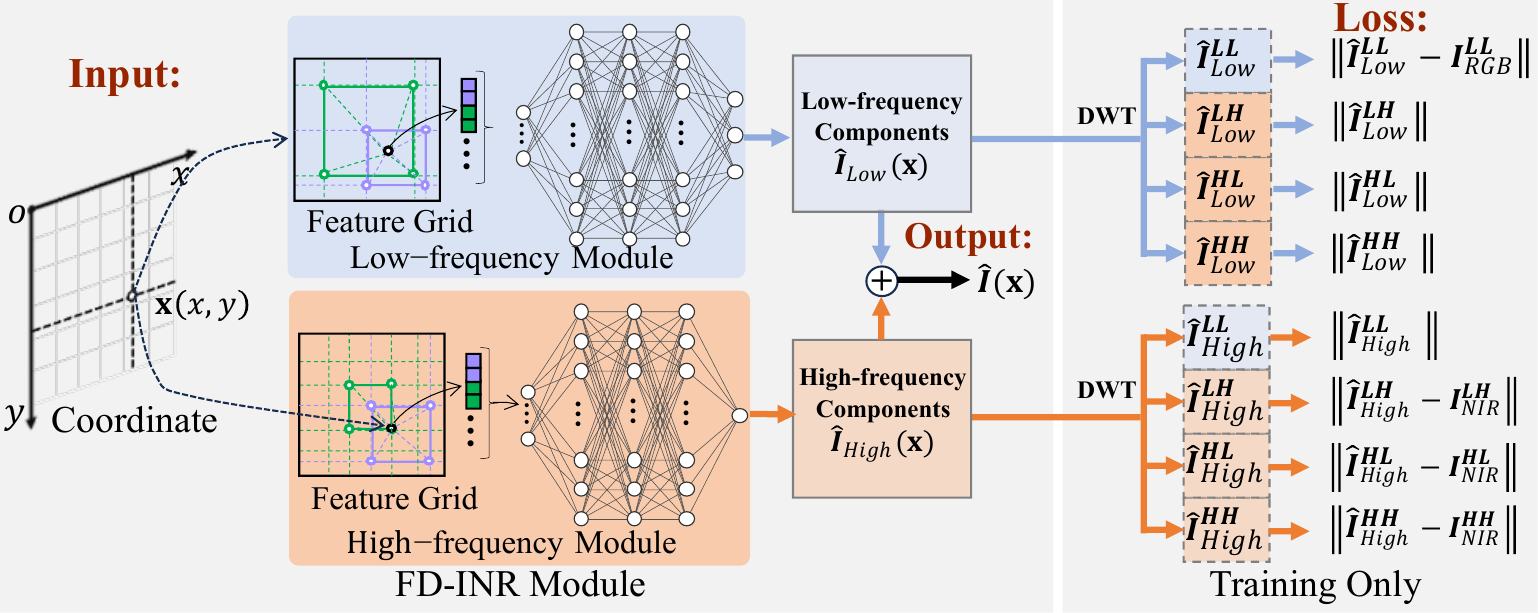}
	\caption{
		\textbf{The architecture of the proposed FD-INR.} Our framework explicitly decomposes the image field into two specialized components: a low-frequency branch ($\mathcal{F}_{LF}$) and a high-frequency branch ($\mathcal{G}_{HF}$). During the optimization, we employ a multi-scale wavelet supervision mechanism where the low-frequency branch is constrained by the color manifold of the RGB input, while the high-frequency branch is guided by the structural priors of the NIR signal.
	}
	\label{fig0}
\end{figure*}

Despite the success of INR in image super-resolution and reconstruction, its potential in low-light enhancement and multi-modal fusion remains under-explored. Existing INR frameworks typically use a single path for all information, which faces Spectral Bias Conflict when processing cross-modal signals like RGB and NIR—a single MLP struggles to fit high-frequency textures and low-frequency colors with distinct statistical distributions under a unified coordinate mapping. To address this, we propose a frequency partitioning strategy. Unlike black-box fusion in latent space, FD-INR decomposes the task into coordinate mapping problems across different frequency bands via explicit wavelet decoupling. This design introduces an inductive bias that reduces the search space for the function, enabling the network to specifically capture NIR structures and RGB global distributions. Furthermore, this architecture inherits the resolution-independent advantages of INR, providing a robust representation for multi-modal perception in extreme environments.

\section{Methodology}\label{sec:methodology}
This section describes the proposed FD-INR, a frequency-decomposed implicit neural representation that achieves cross-modal frequency partitioning within a continuous coordinate space. The core concept leverages varying structural inductive biases of representation primitives to separately parameterize the color manifold of RGB signals and the structural textures of Near-Infrared (NIR) signals. As illustrated in Fig. \ref{fig0}, we construct an asymmetric dual-branch model and introduce a frequency-domain supervision mechanism. This forces the model to extract complementary information within the most physically reliable frequency ranges of each modality, thereby achieving sub-pixel detail restoration while suppressing noise in extremely low signal-to-noise ratio environments.

\subsection{Implicit Representation with Frequency Partitioning}\label{subsec:framework}
We model the image field as a linear superposition of two continuous functions with complementary spectral characteristics. For a given normalized spatial coordinate $\mathbf{x} \in [-1, 1]^2$, the generation process of the reconstructed image field $\hat{\mathbf{I}}(\mathbf{x})$ is defined as:
\begin{equation}
	\hat{\mathbf{I}}(\mathbf{x}) = \mathcal{F}_{LF}(\mathbf{x}; \theta) + \mathcal{G}_{HF}(\mathbf{x}; \phi) - \beta,
\end{equation}
where $\mathcal{F}_{LF}$ denotes the representation field capturing low-frequency color consistency, $\mathcal{G}_{HF}$ represents the field reconstructing high-frequency structural details, and $\beta$ is a learnable bias used to calibrate the Direct Current (DC) component differences between modalities. To implement spectral constraints at the architectural level, we control the spatial sampling resolution of the feature primitives in each branch to guide the model toward specific spectral intervals. Specifically, $\mathcal{F}_{LF}$ utilizes sparse feature grids as primitives. According to sampling theory, sparse sampling naturally limits the maximum frequency the function can represent, resulting in a low-pass characteristic when fitting low-light RGB images, which filters out high-frequency random noise in the continuous space. Conversely, $\mathcal{G}_{HF}$ employs dense feature grids to capture sub-pixel gradients. The high-resolution feature grids allow the model to bypass the "spectral bias" of implicit neural representations, injecting fine structures from NIR images with high fidelity.

\subsection{Wavelet-Domain Spectral Supervision}\label{subsec:supervision}To ensure that the dual branches strictly follow the frequency decoupling prior during optimization, we introduce the Multi-scale Discrete Wavelet Transform (DWT) as a spectral constraint operator. DWT offers excellent time-frequency localization, enabling the precise separation of structural information across multiple scales. Let $\mathcal{W}(\cdot)$ be the wavelet decomposition operator. We project the predicted low-frequency and high-frequency terms into the wavelet domain to obtain sets of coefficients across scales:
\begin{align}
	\{\hat{\mathbf{c}}_{LF}^{(s, \lambda)}\}_{s=1}^{S} &= \mathcal{W}(\mathcal{F}_{LF}), \\
	\{\hat{\mathbf{c}}_{HF}^{(s, \lambda)}\}_{s=1}^{S} &= \mathcal{W}(\mathcal{G}_{HF}),
\end{align}
where $S$ denotes the total number of wavelet decomposition levels (set to $S=3$ in our experiments), and $s \in \{1, \dots, S\}$ is the current decomposition scale. $\lambda \in \{LL, LH, HL, HH\}$ represents the subband index, where $LL$ is the low-frequency approximation subband, and $\Omega_H = \{LH, HL, HH\}$ represents the high-frequency detail subbands in horizontal, vertical, and diagonal directions, respectively. To eliminate feature aliasing between modalities and prevent spectral leakage, we apply Spectral Decoupling Regularization, which constrains the energy of each branch within specific spectral boundaries via forced spectral sparsity:
\begin{equation}
	\mathcal{L}_{reg} = \sum_{s=1}^{S} \left(\sum_{\lambda \in \Omega_H} \|\hat{\mathbf{c}}_{LF}^{(s, \lambda)}\|_1 + \|\hat{\mathbf{c}}_{HF}^{(s, LL)}\|_1 \right).
\end{equation}
In this expression, the first term penalizes the high-frequency subband norms of the low-frequency branch $\mathcal{F}{LF}$ to enforce spatial smoothness and avoid the incorporation of high-frequency noise from the raw RGB image. The second term penalizes the low-frequency approximation of the high-frequency branch $\mathcal{G}{HF}$, ensuring it learns only zero-mean high-frequency details and preventing NIR luminance information from contaminating the RGB colors.
\subsection{Adaptive Optimization and Loss Functions}\label{subsec:loss}
Given the differences in numerical magnitude, convergence speed, and signal reliability between color restoration and structural enhancement tasks, we construct a multi-scale fidelity loss function and introduce an adaptive weighting strategy to balance these tasks.
\paragraph{Low-frequency Color Loss.} We utilize the wavelet approximation subband $\mathbf{T}_{RGB}^{(s, LL)}$ of the low-light RGB image as the supervisory signal. To improve the model's robustness against outlier noise in extreme low-light conditions, we employ the Charbonnier loss:
\begin{equation}\mathcal{L}_{LF} = \sum_{s=1}^{S} \sqrt{| \hat{\mathbf{c}}_{LF}^{(s, LL)} - \mathbf{T}_{RGB}^{(s, LL)} |^2 + \epsilon^2},
\end{equation}
where $\hat{\mathbf{c}}_{LF}^{(s, LL)}$ represents the approximation coefficients of the low-frequency branch at scale $s$, and $\epsilon$ is a tolerance parameter (set to $1 \times 10^{-3}$ in our experiments) to ensure continuous differentiability.
\paragraph{High-frequency Structure Loss.} We use the high-SNR detail subbands $\mathbf{T}_{NIR}^{(s, \omega)}$ of the NIR image for guidance. To balance pixel-level accuracy with structural consistency, we combine $\ell_1$ loss with the Structural Similarity (SSIM) loss:
\begin{equation}
	\begin{split}
		\mathcal{L}_{HF} = \sum_{s=1}^{S} \sum_{\omega \in \Omega_H} \Big( &\lambda_1 \| \hat{\mathbf{c}}_{HF}^{(s, \omega)} - \mathbf{T}_{NIR}^{(s, \omega)} \|_1 \\
		+ &\lambda_2 (1 - \text{SSIM}(\hat{\mathbf{c}}_{HF}^{(s, \omega)}, \mathbf{T}_{NIR}^{(s, \omega)})) \Big),
	\end{split}
\end{equation}
where $\hat{\mathbf{c}}_{HF}^{(s, \omega)}$ denotes the predicted coefficients of the high-frequency branch in subband $\omega \in \Omega_H$. $\lambda_1$ and $\lambda_2$ are the balancing coefficients for the $\ell_1$ and SSIM terms, respectively, both set to $0.5$ in our implementation.
\paragraph{Spatial Gradient Consistency Loss.} To further enhance the edge sharpness of the reconstructed image $\hat{\mathbf{I}}$ and ensure structural consistency with the NIR reference, we introduce a gradient loss:
\begin{equation}
	\mathcal{L}_{grad} = \|\nabla_x \hat{\mathbf{I}}_{Y} - \nabla_x \mathbf{I}_{NIR}\|_1 + \|\nabla_y \hat{\mathbf{I}}_{Y} - \nabla_y \mathbf{I}_{NIR}\|_1,
\end{equation}
where $\hat{\mathbf{I}}_{Y}$ is the luminance component of the reconstructed image (the Y-channel after RGB-to-YCbCr conversion), and $\nabla_x, \nabla_y$ represent the differential operators in the horizontal and vertical directions. This term forces gradient alignment, compensating for the limitations of wavelet-domain supervision in maintaining spatial continuity.
\paragraph{Zero-mean Centering Constraint.} Since the physical function of $\mathcal{G}_{HF}$ is strictly limited to providing sub-pixel high-frequency details, its output field should statistically exhibit zero-mean residual characteristics and should not carry global luminance components. Therefore, we introduce the zero-mean centering constraint
\begin{equation}
	\mathcal{L}_{zero} = \|\text{mean}(\mathcal{G}_{HF} - \beta)\|_2.
\end{equation}
In this expression, the learnable parameter $\beta$ independently absorbs the systemic luminance offset between the RGB and NIR sensors. By minimizing the deviation of $\mathcal{G}_{HF}$ relative to $\beta$, this constraint forces the high-frequency branch to concentrate its representational capacity on local structural fluctuations, effectively preventing "energy leakage"—where strong NIR luminance information erroneously overlays the color field $\mathcal{F}_{LF}$. This mechanism prevents color distortion common in cross-modal fusion, ensuring that the final reconstruction maintains the chromatic fidelity of the original scene while enhancing details.
\paragraph{Adaptive Weighting Strategy} To dynamically balance the aforementioned tasks and handle potential inconsistencies in numerical magnitudes, we incorporate homoscedastic uncertainty modeling. The total loss function is defined as:
\begin{equation}
	\begin{split}
		\mathcal{L}_{total} = &\sum_{k \in \{LF, HF, grad\}} \left( \frac{1}{2\sigma_k^2} \mathcal{L}_k + \log \sigma_k \right) \\
		&+ \alpha \mathcal{L}_{reg} + \gamma \mathcal{L}_{zero},
	\end{split}
\end{equation}
where $\sigma_k$ is the learnable noise variance for task $k$, used to automatically adjust the weight of each loss. $\alpha$ and $\gamma$ are preset hyperparameters. This mechanism adaptively adjusts the contribution ratio of each loss based on the noise levels encountered during training.

\section{Experiments}\label{sec:experiments}
This section evaluates the effectiveness of the proposed FD-INR on cross-modal frequency partitioning tasks using public benchmark datasets. We provide quantitative and qualitative comparisons between FD-INR and representative existing methods, followed by a comprehensive ablation study to analyze the contribution of each core component to the overall performance.
\subsection{Experimental Setup}\label{subsec:setup}
\paragraph{Datasets and Evaluation Protocol}Unlike traditional deep learning-based methods, FD-INR employs instance-specific fitting via implicit neural representations, possessing a distinct \textbf{zero-shot} restoration attribute—meaning the model does not require pre-training on large-scale datasets. We evaluate the performance of FD-INR on the \textbf{Dark Vision Dataset (DVD)} \cite{jin2022darkvisionnet}, a benchmark specifically established for RGB-NIR fusion. To rigorously verify the model's capability in suppressing authentic physical noise (e.g., shot noise and read noise), we prioritize the \textbf{real-world capture subset} for evaluation. This subset consists of \textbf{10 distinct scenes}, with each scene providing a synchronized pair of low-light RGB and NIR images. To preserve the high-fidelity structural features and original sensor characteristics, all images are processed at their native resolution of \textbf{$1880 \times 1040$}.

\paragraph{Architecture Configurations}
To achieve explicit frequency partitioning, both branches of FD-INR are constructed based on multi-resolution feature grids. The core of this design lies in introducing frequency-selective biases at the architectural level by manipulating the spatial sampling rates. For an input image of resolution $H \times W$, the specific configurations are as follows:
\begin{itemize}
	\item \textbf{Low-frequency Branch $\mathcal{F}_{LF}$}: Designed to capture global color consistency. The resolution range of the feature grids is linearly distributed between $[H/16, H/4] \times [W/16, W/4]$. According to sampling theory, such sparse sampling explicitly constrains the representation bandwidth of the function space, creating a structural low-pass filtering effect that naturally suppresses high-frequency color noise during the fitting process.
	\item \textbf{High-frequency Branch $\mathcal{G}_{HF}$}: Focused on sub-pixel structural reconstruction. Its resolution is set between $[H/4, H/2] \times [W/4, W/2]$. By doubling the sampling density relative to the low-frequency branch, this branch effectively mitigates the "spectral bias" common in implicit representations, capturing fine gradients and structural information from the NIR signal with high fidelity.
\end{itemize}
Both branches are followed by a 3-layer MLP (hidden dimension of 128) using ReLU activation.
\paragraph{Implementation Details}
The framework is implemented in PyTorch 2.8.0, and all experiments were conducted on a workstation equipped with a single NVIDIA RTX 3090 (24GB) GPU. We set the total number of iterations to 10,000 for each pair of RGB-NIR input images. For the optimization strategy, we employ a decoupled scheme: the \textbf{Muon optimizer} \cite{mcginnis2025optimizing} is used for the MLP linear layers with a learning rate of $1 \times 10^{-3}$, leveraging its orthogonal update characteristics to accelerate the convergence of coordinate-space feature mapping; the Adam optimizer is used for updating the hash feature grids.Regarding computational efficiency, for a single-frame inference at $512 \times 512$ resolution, FD-INR requires approximately \textbf{17.6 G FLOPs} with a total parameter count of only \textbf{0.3 M}. Benefiting from the efficient optimization of Muon, the model demonstrates significant robustness to initial hyperparameters during the fitting process, eliminating the need for per-scene manual tuning. The average per-iteration inference latency for a single image is approximately \textbf{20.20 ms}.

\subsection{Comparative Analysis} 
\begin{figure*}[!ht]
	\centering
	\includegraphics[width=\textwidth]{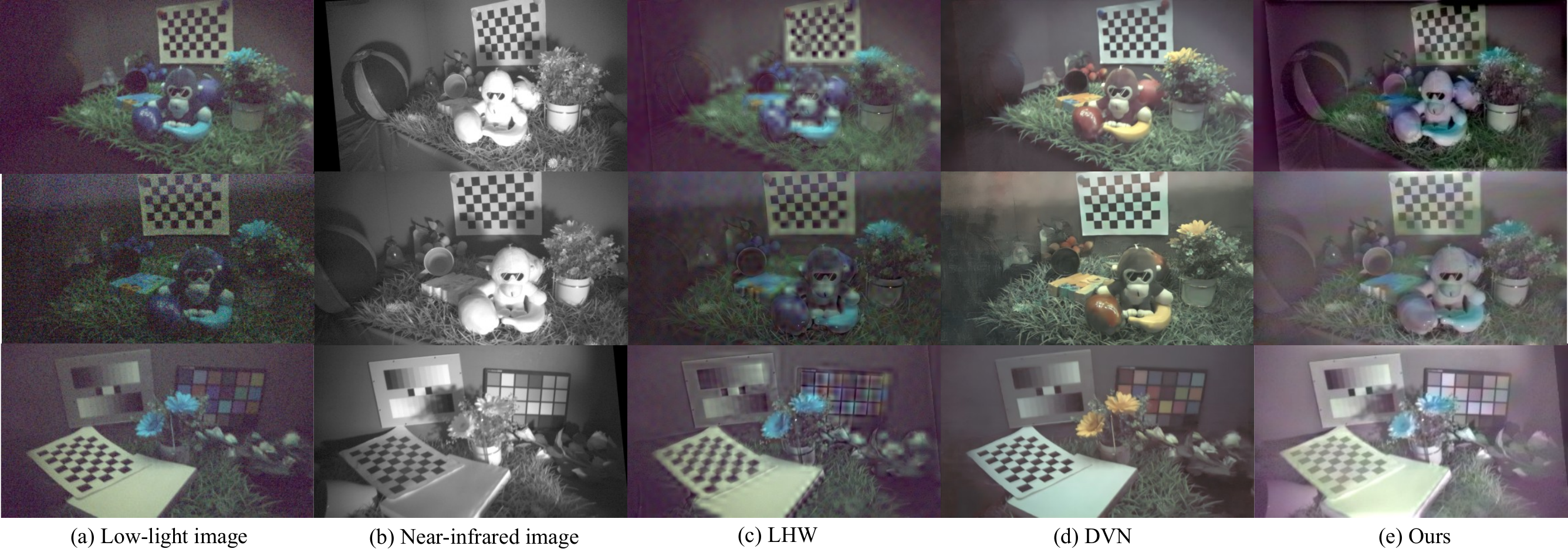}
	\caption{Visual comparison of different fusion paradigms, where (a) Input RGB and (b) NIR (c) Naive DWT-based fusion, (d) DVD baseline method \cite{jin2022darkvisionnet}, and (e) our method. }
	\label{fig1}
\end{figure*}

As visualized in Fig.~\ref{fig1}, we compare our method with two distinct paradigms. (c) represents the Naive DWT Fusion, which suffers from severe color desaturation and spectral leakage due to the lack of irradiance calibration. (d) shows the results of the DVN \cite{jin2022darkvisionnet}, which, while improving brightness, retains visible chrominance noise and exhibits artifacts near sharp boundaries. In contrast, FD-INR (e) produces the most visually pleasing results with sharp edges and pure color manifolds. This demonstrates that explicit frequency decoupling within a continuous coordinate space provides a superior inductive bias for partitioning structural details from stochastic sensor noise.

\subsection{Ablation Study and Analysis} \label{subsec:ablation}

\begin{figure*}[!ht]
	\centering
	\includegraphics[width=\textwidth]{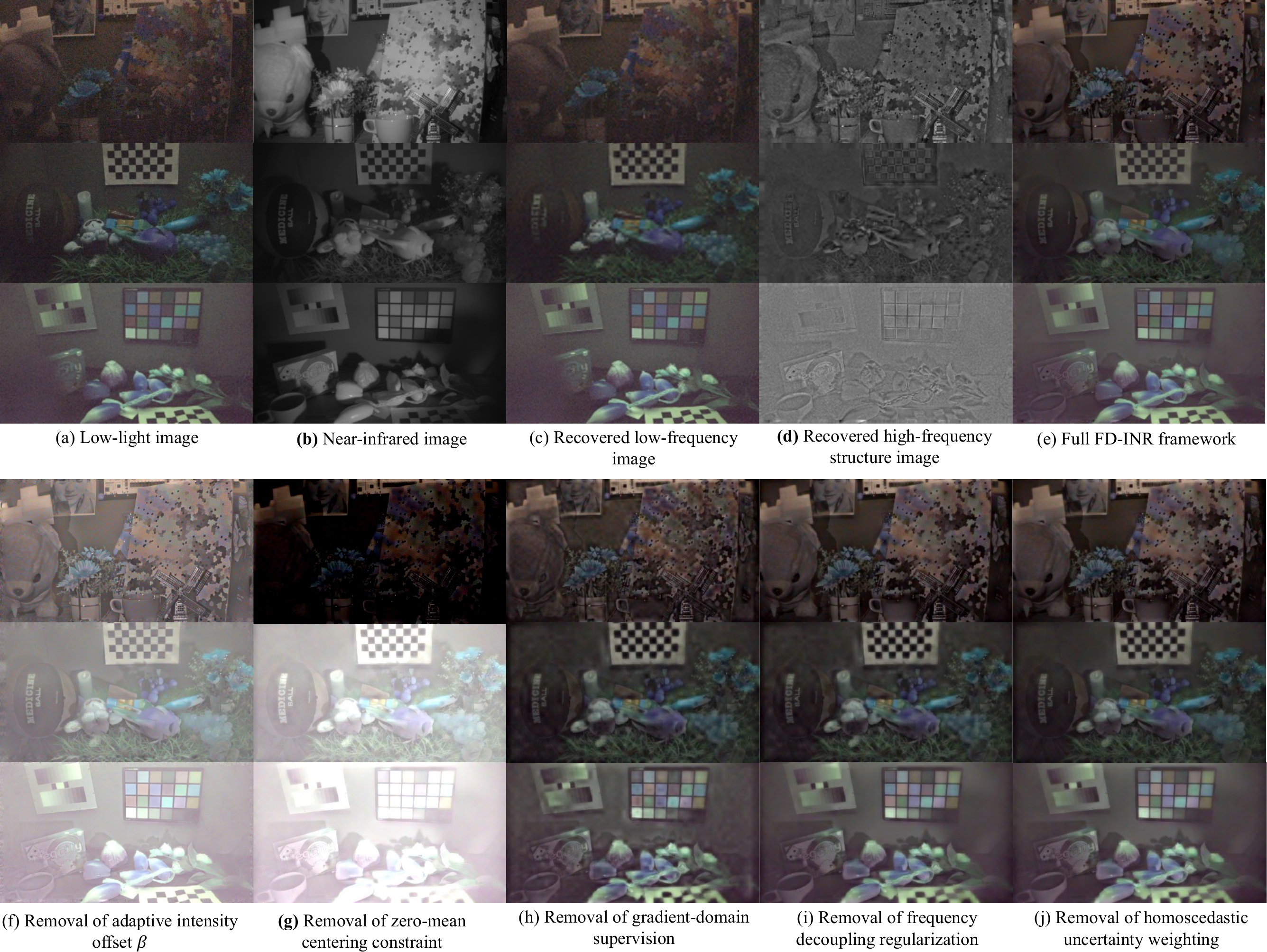}
	\caption{Qualitative ablation study and component visualization of FD-INR. (a)–(e) visualize the explicit spectral decoupling process, where the low-frequency branch captures color manifolds and the high-frequency branch extracts pure textures. (f)–(j) validate the necessity of each component: $\beta$ and the centering constraint prevent overexposure; gradient loss ensures edge sharpness; decoupling regularization prevents spectral overlapping; and uncertainty weighting balances multi-modal constraints.}
	\label{fig2}
\end{figure*}

\begin{table}[ht]
	\centering
	\caption{Quantitative ablation results for the DVD dataset. We report the average \textbf{NIQE} (lower is better) and \textbf{NCC} (higher is better) across representative scenes. All metrics are calculated after cropping 16 pixels from the boundaries to avoid boundary artifacts.}
	\label{tab:ablation_metrics}
	\begin{tabular}{lcc}
		\toprule
		Configuration & NIQE $\downarrow$ & NCC $\uparrow$ \\
		\midrule
		(f) w/o Adaptive Offset $\beta$ & 3.211 & 0.896 \\
		(g) w/o Zero-mean Centering & 3.817 & 0.788 \\
		(h) w/o Gradient-domain Supervision & 3.408 & 0.863 \\
		(i) w/o Frequency Decoupling Reg. & 3.073 & 0.913 \\
		(j) w/o Homoscedastic Uncertainty & 2.994 & 0.915 \\
		\textbf{(e) Full FD-INR (Ours)} & \textbf{2.896} & \textbf{0.961} \\
		\bottomrule
	\end{tabular}
\end{table}

To evaluate the contribution of each component in FD-INR to cross-modal enhancement performance, we conducted a comprehensive ablation study on the real-world DVD dataset. As shown in Table~\ref{tab:ablation_metrics} and Fig.~\ref{fig2}, our analysis covers three dimensions: spectral decoupling, adaptive calibration, and structural regularization.

\paragraph{Visualization of Spectral Decoupling.} 
Fig.~\ref{fig2}(c) and (d) visualize the intermediate outputs of our dual-branch INR. The low-frequency branch $\mathcal{F}_{LF}$ reconstructs a smooth color manifold, effectively suppressing sensor noise via the sparse sampling grid bias. Meanwhile, the high-frequency branch $\mathcal{G}_{HF}$ (Fig.~\ref{fig2}(d)) extracts sharp structural details under NIR guidance. This explicit decoupling ensures that chromatic information and structural features are optimized within their respective specialized coordinate spaces.

\paragraph{Impact of Adaptive Calibration ($\beta$ and Centering).} 
The primary challenge in RGB-NIR fusion is the irradiance mismatch. As shown in columns (f) and (g), the absence of offset calibration leads to severe overexposure and color desaturation. Our zero-mean centering constraint ensures that the high-frequency branch functions strictly as a residual texture field, maintaining the physical energy balance between modalities. Notably, without the centering constraint, the reconstruction results exhibit extreme overexposure or unnatural darkening (see column (g), rows 2 and 3).

\paragraph{Effectiveness of Decoupling Regularization $L_{reg}$.} 
Column (i) presents the results when frequency decoupling regularization is removed. Specifically, in the second row of column (i), the checkerboard pattern is significantly blurred compared to the full model [Fig.~\ref{fig2}(e)]. This indicates that removing $L_{reg}$ causes spectral overlapping, where high-frequency structural details are smeared by low-frequency color leakage, failing to exploit the high-resolution potential of the NIR guide.

\paragraph{Gradient Supervision and Uncertainty Weighting.} 
The results in column (h) display a noticeable "oil-painting" effect with blurred edges and lost fine textures, proving that gradient-domain supervision is critical to overcome the over-smoothing bias inherent in coordinate-based MLPs. Finally, column (j) demonstrates the robustness of homoscedastic uncertainty weighting ($u\_loss$) in balancing multi-modal constraints. Without this mechanism, the model struggles to weight RGB consistency against NIR structural guidance, leading to unstable performance in regions with extreme sensor noise.

\paragraph{Quantitative Analysis.}
As summarized in Table~\ref{tab:ablation_metrics}, the quantitative metrics align consistently with our qualitative observations. The full FD-INR model achieves superior performance in both perceptual naturalness (lowest NIQE) and structural fidelity (highest NCC). Specifically, the removal of the centering constraint (g) significantly degrades NIQE, quantifying the distortion caused by irradiance mismatch, while the absence of gradient supervision (h) directly leads to a decrease in NCC, reflecting the loss of edge sharpness.

\section{Conclusion}
To address the challenges of extremely low signal-to-noise ratios and cross-modal representation conflicts in low-light imaging, this paper proposes FD-INR, a frequency-decoupled implicit neural representation framework. From the perspective of continuous function space, our approach utilizes sampling theory to construct structured inductive biases, aiming to model the physical prior where the RGB modality establishes the color manifold and the NIR modality enhances high-frequency details. Through an asymmetric dual-branch implicit architecture combined with wavelet-domain constraints, FD-INR provides a feasible solution for partitioning noise from texture. Furthermore, the application of an adaptive uncertainty-based weighting mechanism helps balance the contributions of multi-modal data across different spectral bands and alleviates color distortion issues during the fusion process to a certain extent. Experimental results indicate that FD-INR yields competitive restoration performance across multiple benchmark datasets and demonstrates potential for resolution-agnostic reconstruction, leveraging the continuous nature of implicit neural representations.

\section*{Impact Statement}
This paper presents work whose goal is to advance the field of Machine
Learning. There are many potential societal consequences of our work, none
which we feel must be specifically highlighted here.
\nocite{langley00}

\bibliography{example_paper}
\bibliographystyle{icml2026}

\end{document}